\pdfoutput=1

\documentclass[twocolumn,apacite,authoryear]{els-mrw} 
\usepackage{amsmath,amssymb,amsfonts,amsthm,makeidx,graphicx}
\usepackage{txfonts}
\usepackage{helvet}

\renewcommand{\articlecopyright}{\vspace{-6pt}}

\begin{document}

\chapter{Natural Language Generation}\label{chap1}

\author[1]{Emiel van Miltenburg}%
\author[2]{Chenghua Lin}%

\address[1]{\orgname{Tilburg University}, \orgdiv{Department of Cognition and Communication}, \orgaddress{Tilburg, The Netherlands}}
\address[2]{\orgname{University of Manchester}, \orgdiv{Department of Computer Science}, \orgaddress{Manchester, UK}}

%\articletag{Chapter Article tagline: update of previous edition,, reprint..}

\maketitle
\begin{abstract}[Abstract]
Natural Language Generation (NLG) regards the automatic production of text. This overview focuses on common applications, approaches to text generation, and ways to evaluate NLG systems. A list of open research questions is also provided. 
\end{abstract}

\section*{Keywords}
Natural Language Processing; Language production; Large Language Models; Applications; Evaluation

\section*{Key points}
\begin{itemize}
    \item The field of Natural Language Generation is dedicated to the automatic generation of text. Classical applications in business, journalism, and medicine involve data-to-text generation.
    \item Traditional NLG follows a pipeline approach that is typically reliable but less fluent, while modern NLG often uses LLMs which are more fluent, but less reliable.
    \item Evaluation of output quality is either done through human evaluation (still the gold standard) or through automatic means (more efficient, but may have a lower construct validity).
    \item The main challenge in NLG is how to generate text that is both fluent \emph{and} factually correct, particularly for longer texts, where \emph{coherence} is also hard to achieve. Performing reproducible and high-quality evaluations remains difficult.
    \item Given the many applications of NLG technology, researchers should be aware of the social implications of their work.
\end{itemize}

\section{Introduction}
The term \textit{Natural Language Generation} (NLG), in its broadest definition, refers to the study of systems that verbalize some form of information through natural language.\footnote{Either unconstrained or a  \emph{Controlled Natural Language} (CNL; see \citealt{kuhn-2014-survey}).} That information could be stored in a large database or knowledge graph (in \emph{data-to-text} applications), but NLG researchers may also study summarisation (\emph{text-to-text}) or image captioning (\emph{image-to-text}), for example. As a subfield of Natural Language Processing, NLG is closely related to other sub-disciplines such as Machine Translation (MT) and Dialog Systems. Some NLG researchers exclude MT from their definition of the field, since there is no \emph{content selection} involved where the system has to determine \emph{what to say}. Conversely, \emph{dialog systems} do not typically fall under the header of Natural Language Generation since NLG is just one component of dialog systems (the others being Natural Language Understanding and Dialog Management). However, with the rise of Large Language Models (LLMs), different subfields of Natural Language Processing have converged on similar methodologies for the production of natural language and the evaluation of automatically generated text.

\subsection{Relation to linguistics}
Historically, NLG research has had close ties with pragmatics and psycholinguistics. The reason for this is that automatically generated text should generally be as fluent and natural as possible. However, there are often many different ways to communicate the same message. How to best formulate that message depends on the context.\footnote{We can perhaps best see this in the literature on referring expression generation \citep{krahmer-van-deemter-2012-computational}, but other areas of NLG research also reference the pragmatics literature, particularly the Gricean maxims (\citealt{grice1975logic}; see \citealt{krause-vossen-2024-gricean-maxims} for a survey of the use of the maxims in NLG).} By studying human language production, we can better understand how people formulate their utterances in different situations. Psycholinguistic models of speech production (e.g.\ \citealt{Levelt1989}) also have clear parallels with classical NLG pipelines (see \S\ref{sec:classical}). All these models can be divided into two stages: conceptualisation (\emph{what to say}) and formulation (\emph{how to say it}).

\subsection{Applications}
NLG technology promises to automate or streamline writing tasks in many different domains. We discuss three examples below.

\subsubsection{NLG and business}
The commercial potential of NLG has long been recognized; the first company to commercialize NLG was founded in 1990 and several different companies have followed in their footsteps \citep{Dale_2020}. Robert \citeauthor{Dale_2020} marks 2012 as the year when NLG entered the mainstream media conscious, with the publication of a \textit{Wired} article on automated journalism. Following the release of ChatGPT, a full decade later (November 30, 2022), we have now seen how automatic text generation has become fully mainstream, with different companies offering the general public paid subscriptions to Large Language Models via user-friendly interfaces.

NLG techniques have traditionally been used to convert tabular data into text. For example, generating financial reports,\footnote{Ehud \citet{reiter2018commercial} notes that all major NLG companies are involved in financial reporting, because the sector is large and use cases are similar between different organisations, making it easier to capitalise on NLG technology than in areas where more customisation is needed.} or generating product titles and descriptions based on product specifications \citep{mathur-etal-2017-generating,zou2023automatic}. Following the rise of ChatGPT, we are now seeing an explosion of interest to use language models for all sorts of purposes, including content generation (writing informative texts), and the generation of email responses (see \citealt{tafesse2024chatgpt,tafesse2024hey} for more discussion).

\subsubsection{NLG and journalism}
Like with business applications, NLG techniques have traditionally been used in journalism to produce data-driven articles to report on topics such as weather reports, sports matches, earthquakes, and elections. These articles are relatively formulaic, making it easier to develop templates where relevant values (such as the magnitude of an earthquake) can be inserted \citep{gatt2018survey}. Nowadays, journalists use generative AI throughout the reporting process (news gathering, news production, news verification, distribution, and moderation) for a much wider array of articles \citep{Cools26082024}. Still, classical rule- and template-based approaches remain the most reliable solution for standalone text generation, since they do not suffer from hallucination (see \S\ref{sec:factuality}). Moreover, as \citet[Chapter 6]{Reiter2025} notes, rule-based systems may be easier to modify (e.g.\ make small changes to the output) and maintain.

\subsubsection{NLG in a medical context}
There is a long history of Natural Language Generation in healthcare (see \citealt{10.1136/jamia.1997.0040473} for an early survey). Applications range from medical report generation (presenting relevant clinical data to different audiences, e.g.\ \citealt{gatt2009data}) to clinical note generation (summarising doctor-patient interactions; e.g.\ \citealt{ben-abacha-etal-2023-investigation}) and personalised decision support tools (e.g.\ \citealt{hommes-etal-2019-personalized}). Challenges in the medical field include working with potentially unreliable sensor data \citep{10.1093/oxfordhb/9780198736578.013.32}, and involving different stakeholders (doctors, patients, nurses) who often differ in background knowledge, reading level, technical ability, and socio-economic status. Medical NLG research is usually carried out in interdisciplinary teams, with experts not only assessing the quality of the output, but also looking at the potentially harmful impact that systems may have on their users (e.g.\ \citealt{balloccu-etal-2024-ask}).

\section{Approaches}
This section provides a brief overview of the different approaches to Natural Language Generation, focusing on the evolution from the classical NLG pipeline, through end-to-end approaches, to the use of (more general) large language models. (For an overview of all NLG approaches before LLMs, see \citealt{gatt2018survey}). 

\subsection{The classical NLG pipeline}\label{sec:classical}
In the classical NLG pipeline, text generation proceeds in different stages. Here are the steps proposed by \citet{Reiter_Dale_2000}:
\begin{itemize}
    \item \textbf{Content determination}: deciding what information should be communicated.
    \item \textbf{Document structuring}: deciding how to group and order different chunks of information.
    \item \textbf{Lexicalisation}: deciding what words, phrases, or syntactic constructions to use.
    \item \textbf{Referring expression generation}: deciding what words or phrases should be used for different named entities.
    \item \textbf{Aggregation}: deciding how to map the different information chunks to sentences and paragraphs.
    \item \textbf{Realisation}: converting the abstract representation of the text-to-be-generated into a real text.
\end{itemize}

These steps can be handled by different modules that may either use (often hand-written) rules, Machine Learning, or both. The initial steps of the pipeline are context-dependent, but the final stage (realisation) is universal. As such, different projects may re-use the same \emph{realiser}.\footnote{Among existing realisers, \textsc{SimpleNLG} \citep{gatt-reiter-2009-simplenlg} is a popular choice that has been translated into several different languages.}

\subsection{End-to-end generation}
End-to-end generation drastically simplifies the problem of natural language generation by directly mapping data to text. All steps are implicitly handled by the same model that is developed using a data-driven approach (see \citealt{castro-ferreira-etal-2019-neural} for a discussion). Popular challenges that can be resolved through this approach are the E2E dataset \citep{novikova-etal-2017-e2e} and the WebNLG challenge (e.g.\ \citealt{cripwell-etal-2023-2023}).\footnote{Although note that these challenges do not require any signal analysis or content determination.} Taking WebNLG as an example, the goal is to render a set of RDF triples in natural language. Recent years have seen the focus shift away from English (as this problem has more or less been solved) and towards low-resource languages.

\subsection{Large Language Models}
Large Language Models (LLMs), such as ChatGPT \citep{ouyang2022training} and LLaMA \citep{touvron2023llama}, typically adopt transformer-based architectures \citep{vaswani2017attention}, which rely on self-attention mechanisms to capture long-range contextual dependencies in text. The training of LLMs generally follows a two-stage training pipeline.
The first stage, known as \textit{pre-training}, is an unsupervised process in which models are trained on vast amounts of unlabelled textual data containing billions or even trillions of tokens from diverse sources such as web pages, books, and encyclopaedias. This stage allows LLMs to acquire and encode general linguistic and world knowledge.
The second stage, known as \textit{Supervised Fine-Tuning} (SFT), involves further training the model on labelled datasets containing explicit instruction-output pairs. SFT aims to enhance the model's ability to understand and follow human-provided instructions, ensuring its responses are more relevant, specific, and aligned with user expectations. A notable variant of this stage is instruction tuning \citep{wei2022finetuned}, which expands the fine-tuning process to a diverse set of tasks expressed as natural language instructions, thereby improving generalisation to unseen prompts.
Techniques such as reinforcement learning from human feedback (RLHF) and direct preference optimisation (DPO) can furether refine model behaviour, enhancing safety, helpfulness, and alignment with human intent \citep{10.5555/3692070.3694333}. Extending these approaches, multi-objective alignment (e.g., MetaAligner; \citealt{yang2024metaaligner}) and mixture-of-preference optimisation (e.g., MODPO; \citealt{zhou-etal-2024-beyond}) seek to balance competing objectives (e.g. helpfulness versus harmlessness), while maintaining performance across targets.

%The second stage, known as \textit{Supervised Fine-Tuning} (SFT), involves further training the model on labelled datasets containing explicit instruction-output pairs. SFT aims to enhance the model's ability to understand and follow human-provided instructions, ensuring its responses are more relevant, specific, and aligned with user expectations. Techniques such as reinforcement learning from human feedback (RLHF) and direct preference optimisation (DPO) can further refine the model’s outputs, improving its safety, helpfulness, and alignment with human preferences \citep{10.5555/3692070.3694333}. 

\subsubsection{Prompting}
Once trained, LLMs generate text through prompting, where users provide textual inputs that may include explicit instructions, context, or examples to guide the model’s responses. Prompting enables LLMs to perform a wide range of tasks, such as summarisation, translation, question answering, and creative writing, without requiring additional task-specific fine-tuning. Advanced prompting techniques, such as chain-of-thought prompting that introduces intermediate reasoning steps within the prompt, can further enhance the model’s performance on complex reasoning and generation tasks (see \citet{10.1145/3560815} and \citet{schulhoff2025promptreportsystematicsurvey} for recent surveys on this topic). However, LLMs are highly sensitive to prompt design; even minor differences between prompts can cause substantial differences in output quality \citep{DBLP:conf/iclr/Sclar0TS24}. This phenomenon is commonly referred to as \emph{prompt brittleness} (see e.g. \citealt{gao2025take} for discussion).

\subsubsection{Prompt optimisation}
The brittleness of LLMs using human-authored prompts has motivated the development of the subfield of Automatic Prompt Optimization (APO). Current approaches to APO primarily fall into three categories: gradient-based optimization, iterative search, and inversion-based prompting. 
Gradient-based methods represent prompts as continuous vectors that can be optimized through gradient descent~\citep{pryzant-etal-2023-automatic}. Search-based methods, by contrast, iteratively refine candidate prompts based on performance on a validation set~\citep{wen-etal-2025-hpss}. Inversion-based methods adopt a generative perspective, learning reverse mappings from model outputs to input instructions to automatically generate effective, model-specific prompts~\citep{hong2025beyond}.

% TODO: Needs to have something on automatic methods to optimize prompts

\subsection{Hybrid approaches: pipelines and LLMs}
The distinctions made above are not absolute: we may also see \emph{hybrid} approaches to Natural Language Generations where, for example, different parts of the pipeline are replaced by LLMs or end-to-end models. Indeed, earlier work by \citet{castro-ferreira-etal-2019-neural} has shown that having explicit intermediate steps (as in the pipeline approach) may be beneficial to the output quality of end-to-end models. Hybrid approaches may also be easier to control and offer more transparency than a fully LLM-based approach, while still enjoying the benefit of increased output fluency (see also \S\ref{sec:factuality} below). Another hybrid technique is to only rely on LLM outputs if certain conditions are met, and else to fall back on rule-based or predefined responses (e.g.\ \citealt{10.1145/3571884.3604404}).

\section{Evaluation}
Evaluating the quality of automatically generated text remains a fundamental challenge in natural language processing (NLP). As text generation models continue to improve, the difficulty of evaluating their outputs has also increased. In many cases, evaluating generated text is now as challenging as generating it. This difficulty arises due to the inherent variability in natural language—multiple outputs can be equally valid despite differing in lexical choice, structure, or even semantics. For instance, in machine translation, multiple translations of a given sentence may convey the same meaning while varying in word choice and phrasing. The challenge becomes even more pronounced in open-domain dialogue generation, where a single input may lead to many plausible responses with different semantics, i.e. the ``one-to-many'' problem \citep{zhao-etal-2023-evaluating,zhao-etal-2024-slide}. Similarly, summarisation often involves subjectivity, as multiple summaries can be correct depending on the focus and interpretation of the content. Additionally, certain text generation tasks, such as humour or metaphor, require evaluators to account for demographic and cultural differences, further complicating the assessment process \citep{loakman-etal-2023-iron,wang-etal-2024-mmte}. We can rougly distinguish two kinds of evaluation: using human evaluation studies, or through automatic metrics (see \citealt{DBLP:journals/corr/abs-2006-14799} for a survey).

\subsection{Human evaluation}\label{sec:humeval}
In human evaluation studies, participants answer questions about one or more texts \citep{VANDERLEE2021101151}. A common approach is to rate the quality of generated texts on a five-point scale \citep{amidei-etal-2019-use}. There is a wide array of different dimensions that may be relevant to assess the quality of a text, for example: \textit{fluency, correctness, completeness}, and \textit{appropriateness} (see \citealt{belz-etal-2020-disentangling} for a general taxonomy). For quality dimensions that relate to intrinsic properties of the texts, human evaluation studies are similar to traditional experiments that one may also encounter in other subfields of linguistics. For quality dimensions that are more closely related to the use of an NLG system in context, techniques from the field of human-computer interaction (such as contextual interviews, think-aloud tasks, or usability surveys) may be used. Finally, authors may also choose to carry out an \emph{error analysis}, where authors identify and categorise instances where something went wrong in the generated text \citep{van-miltenburg-etal-2021-underreporting}.

\subsection{Automatic evaluation}

While human evaluation remains the gold standard in NLG research, it can be costly and time-consuming. Therefore, different researchers have developed different ways to assess the output of NLG systems along different dimensions (e.g. \emph{fluency, grammaticality, correctness}). \citet{DBLP:journals/corr/abs-2006-14799} distinguish two kinds of automatic evaluation:
\begin{enumerate}
    \item \textbf{Untrained automatic metrics} compare generated texts with a set of reference texts, through different similarity measures.
    \item \textbf{Machine-learned metrics} are based on machine-learned models, and serve as automatic equivalents to human judges.
\end{enumerate}

Recent years have seen rapid developments in the second category, with the \emph{LLM-as-a-judge} paradigm quickly gaining popularity after showing impressive results on different benchmarks (e.g.\ \citealt{10.5555/3666122.3668142}). For an overview of currently used tasks and evaluation measures, readers are referred to the recurring GEM shared task (e.g.\ \citealt{mille-etal-2024-2024}), which is a comprehensive community effort to assess NLG models on a wide array of tasks.\footnote{For an indication of the popularity of current metrics, see \citealt{schmidtova-etal-2024-automatic-metrics}. But note that single metrics are always reductive; no individual metric can fully capture all different quality dimensions.}

\section{Challenges}

\subsection{Factuality versus fluency?}\label{sec:factuality}
There is an interesting tension between factuality and fluency of automatically generated texts. Generally speaking, traditional rule-based approaches result in texts that are factually correct but not always fluent, while neural approaches tend to produce texts that are fluent but not always factually correct.\footnote{Though see \citealt{10.1093/oxfordhb/9780198736578.013.32}, where the authors show how all types of NLG systems may deviate from the truth in their outputs. To some extent this is unavoidable in situations where the system also has to carry out signal analysis (i.e.\ interpreting the input, with a risk of misinterpretation) and content selection (meaning that the system cannot provide `the whole truth').} This is related to the problem of hallucination (see \citealt{10.1145/3703155} for a survey). It is currently unclear how to guarantee that NLG output is both fluent \emph{and} 100\% factual.

\subsection{Long Text Generation}
%We are starting to see that current models are capable of generating high-quality shorter texts, but longer texts remain a challenge. 

Traditionally, most text generation tasks have focused on producing relatively short outputs, such as weather reports or summaries that span a few dozen to a few hundred words. However, there is growing interest in developing models capable of generating much longer texts, often extending to thousands of words. 
For example, recent efforts such as AI Scientist \citep{lu2024aiscientist} have demonstrated the ability to generate entire scientific papers. By incorporating structured scientific knowledge (e.g., experimental results), this framework can draft papers that adhere to domain-specific requirements, integrating relevant citations and conforming to disciplinary norms. 
Similarly, LongWriter \citep{bai2024longwriter} addresses long-text generation across various domains, including academic publications and monographs. It employs hierarchical attention mechanisms and fine-tuning strategies to maintain thematic consistency and produce structured arguments across extended outputs. 
Despite notable advancements, achieving coherence and structure in long-form content generation remains a significant challenge. This challenge stems from key issues such as the scarcity of high-quality long-form text data in the supervised fine-tuning phase, which limits the model's ability to learn effective writing patterns. Moreover, the alignment phase based on RLHF or DPO introduces complexities in reward modeling and evaluation, where assessing attributes like narrative flow, logical consistency, and thematic coherence becomes increasingly difficult as the text lengthens.

%ntuitively, this is because longer texts require more effort to monitor what has been said before to remain consistent and avoid repetition.

\subsection{Evaluation}
NLG evaluation is a hotly debated topic. Earlier studies found that there is a widespread confusion about the terminology that is used to refer to different quality dimensions \citep{howcroft-etal-2020-twenty}. Human evaluations may always not be reproducible \citep{belz-etal-2023-non}, which may be partly due to past reporting standards \citep{howcroft-etal-2020-twenty,shimorina-belz-2022-human,Gehrmann2023}, but we also have to consider the inherent difficulty of quantifying the objective quality of a text (if such a thing even exists). More research is needed to deepen our understanding of (the interaction between) different quality dimensions that are involved in the process of text interpretation and assessment \citep{van-miltenburg-etal-2020-gradations}.

\subsection{Reproducibility}
An open challenge in NLG research is how to ensure reproducible results. In other words: being able to obtain the same results as those reported in earlier studies. Although this holds for both the output generated by NLG software and the results of the evaluation procedure, most attention has gone to the repeatability of human evaluation studies \citep{belz-etal-2023-non}.

\subsection{Open versus closed models}
% Models are open-weight, but not fully open.
% Top performing models are locked behind an API.
% Criticisms.
When using an LLM for natural language generation, what model should you choose? There are several state-of-the-art commercial models that are only available through an API, but that seem to perform better than more open models. For scientific work, the case for open models seems quite clear: \citet{rogers2023closed} argue that `closed models make bad baselines' because we know too little about their internals and their training data to have a reasonable scientific discussion about their merits and shortcomings. But openness is not a unidimensional construct: \citet{10.1145/3630106.3659005} discuss fourteen different dimensions of openness, and show that different so-called `open' models differ greatly in the extent to which they are actually open. NLG researchers have to carefully weigh the different properties of the available LLMs before selecting their model of choice.

\subsection{Ethics: the social impact of NLG}
As NLG technology is both increasingly powerful \emph{and} increasingly widespread, we also have to contend with the real-world implications of our work. A recent survey by \citet{vanmiltenburg2025dualuseissuesfield} provides an overview of dual use issues that can arise from our research, and \citet{solaiman2024evaluatingsocialimpactgenerative} offer a broad taxonomy of areas that may be impacted by generative AI systems. (Also see earlier work by \citealt{10.1145/3531146.3533088}.) Handling these issues requires continuous effort from both individual researchers and the community as a whole.

\section{Conclusion}
This article provided an overview of the field of Natural Language Generation. Due to space constraints, we have focused on the application side of the field (i.e.\ building and evaluating systems that convert data to text), rather than on cognitive aspects. NLG researchers have always found inspiration in the way humans produce language, so although we have not focused on this topic, we do encourage readers to also explore the more cognitively oriented work that studies human language production in context (see, for example, the entry on spoken language production in this encyclopedia) and work that compares human and automatic language production. Readers who would like to learn more about Natural Language Generation can read the survey by \citet{gatt2018survey} or the recent book by \citet{Reiter2025}.

\begin{ack}[Acknowledgments]
\ldots
\end{ack}

\seealso{article title article title}\\

\section*{Bibliography}
\bibliographystyle{APA-Style}
\bibliography{reference}

\end{document}